# Entity Recognition from Colloquial Text

Tamara Babaian, Jennifer Xu[1]

*Department of Computer Information Systems, Bentley University, USA*

**Abstract.** Extraction of concepts and entities of interest from non- formal texts such as social media posts and informal communication is an important capability for decision support systems in many domains, including healthcare, customer relationship management, and others. Despite the recent advances in training large language models for a variety of natural language processing tasks, the developed models and techniques have mainly focused on formal texts and do not perform as well on colloquial data, which is characterized by a number of distinct challenges. In our research, we focus on the healthcare domain and investigate the problem of symptom recognition from colloquial texts by designing and evaluating several training strategies for BERT-based model fine-tuning. These strategies are distinguished by the choice of the base model, the training corpora, and application of term perturbations in the training data. The best-performing models trained using these strategies outperform the state-of-the-art specialized symptom recognizer by a large margin. Through a series of experiments, we have found specific patterns of model behavior associated with the training strategies we designed. We present design principles for training strategies for effective entity recognition in colloquial texts based on our findings.

## 1. Introduction

In many application domains, important information needs to be extracted from non-formal, colloquial texts, such as complaints about product defects in customer phone calls [15], software bugs reported by developers or users [72], work-related issues noted in meeting minutes, Zoom transcripts, and emails [5], signs of emotional distresses expressed in social media posts [9, 30], and names, concepts, and events mentioned in human-chatbot dialogs [7]. Such extracted information can be categorized, aggregated, and analyzed to support decision-making in organizations to improve products, services, and/or customer relationships.

Development of large language models (LLMs), such as BERT (Bidirectional Encoder Representations from Transformers) [13] and GPT (Generative Pretrained Transformer) [48], have brought unprecedented potential for improving information retrieval, supporting decision-making, and driving enterprise value. These LLMs could be used to discover invaluable insights into and intelligence about various domains. However, information extraction from colloquial text remains a challenge [56, 67, 69].

This research is intended to develop design principles and training strategies for fine-tuning BERT and its variants for entity and concept recognition from colloquial text. We select BERT-based models because they have shown outstanding performance in entity recognition tasks [16, 40, 42, 66], which remain a challenge for GPT-based models [47], yet BERT-based models require considerably less computational resources [69]. We use the healthcare as the test domain, where there exists a large volume of non-formal, colloquial content, such as narratives and stories posted by patients on social media platforms and transcripts of telemedicine visits. Important entities and concepts, such as diseases, treatments, medicines, and symptoms, can be extracted from these texts to serve a variety of purposes [27, 65]. Specifically, recognizing symptom mentions from these text data may help healthcare professionals and disease control organizations gather timely information and reports from patients and the public. The collected information can be very useful for critical, time-sensitive decision-making [33], such as understanding patient health conditions and reactions to treatments, discovering and documenting side effects of drugs, monitoring the spread of new viruses (e.g., COVID-19), and studying the characteristics and impacts of large-scale epidemics. When incorporated in electronic health record (EHR) systems, symptom recognition from doctor-patient communication can help medical professionals identify, recall, and compare relevant aspects of medical histories of patients, thereby improving quality of care and patient satisfaction.

[1] E-mail addresses: tbabaian@bentley.edu (T. Babaian), jxu@bentley.edu (J. Xu).
Published in Direct Decision Support Systems (www.elsevier.com/locate/dss )

Recognizing symptoms from text is more challenging than recognizing other medical entities even with the help of LLMs [6]. First, unlike other types of medical concepts (e.g., diseases, treatments, and medicines), which have well-defined names and regular patterns, symptoms often are subjective descriptions of patient experiences and feelings [29, 70] and take varying linguistic forms [43]. A symptom can be a noun (e.g., *headache*), an adjective (e.g., *painful, dizzy*), a verb (e.g., *vomit*), a phrase (e.g., *throw up, hard to breath*), or an entire sentence (e.g., "*I feel like I submerged 1000 feet underwater*"). There is no standard vocabulary or ontology for symptoms [40]. Second, the boundary of a symptom description can be ambiguous when the body location (e.g., *pain in the temple*), severity (e.g., *excruciating pain*), or frequency (e.g., *constant pain*) [12, 53] are also mentioned. Merely recognizing the symptom itself (e.g., *pain*) is not sufficient in these situations.

Colloquial content generated by non-professionals poses additional challenges. For instance, unlike formal academic papers or clinical notes, social media posts and patient narratives do not necessarily use medical terminology for symptoms (e.g., *dyspnea* for shortness of breath). Patients may use figurative expressions to describe their symptoms. Social media messages often contain abbreviations, typos, spelling errors, and other styling variants, which add noise to the data.

Finally, there is a lack of high-quality labeled corpora for training language models for symptom recognition [58]. Although a few well-known medical corpora (e.g., MIMIC II/III and i2b2/n2c2) contain clinical notes with some labeled clinical concepts (e.g., diseases and medical procedures) [6], they have only a small number of symptom labels. Moreover, only a limited amount of research on symptom recognition can be found in the literature. Prior studies have trained models using medical papers or clinical notes written by researchers or healthcare professionals [31, 40]. The resulting models tend to perform poorly when used to identify symptoms in informal text [44].

Using design science methodology [23], we investigate performance of BERT and a set of its variants for recognizing symptoms from colloquial writings such as patient narratives posted on social media sites. We seek to address a key research question: *How can these language models be fine-tuned to recognize symptoms from colloquial text effectively?* To address this question, we design and experiment with several training strategies for fine-tuning the models and investigate how different design choices affect model performance. These design strategies can be easily implemented in practice without requiring a large amount of GPUs and storage. Our research is among the first few studies that focus on entity recognition, particularly symptom recognition, from colloquial text.

The remainder of this paper is organized as follows. The next section reviews the literature on symptom recognition and training strategies of language models. Section 3 presents our research design. Section 4 describes data and methods, followed by analysis of results in Section 5. Section 6 presents the summary of findings, recommends several design guidelines, and discusses the implications of this research for literature and practice. The last section concludes the paper.

## 2. Literature Review

### 2.1 Symptom Recognition Approaches

Symptom recognition can be viewed as a type of Named Entity Recognition (NER). NER is a subtask of Natural Language Processing (NLP) that involves identifying entities of interest, such as persons (e.g., John Doe), organizations (e.g., the National Science Foundation), locations (e.g., Boston), date and time (e.g., July 4th, Saturday morning), emails, and URLs (e.g., support@gmail.com, www.nsf.org), from text. Traditional NER methods, including lexicon-based pattern recognition, rule-based heuristics, statistical models, and recurrent neural networks, have been used for symptom recognition [6, 29]. For instance, CLAMP, a toolkit for building clinical NLP pipelines, uses a combination of lexicons based on UMLS (Unified Medical Language System), regular expressions, and CRF (Conditional Random Field) to recognize clinical concepts from text [57]. The CLAMP pipeline has been adopted in a state-of-the-art symptom recognition system, SignSym, to identify COVID related symptoms [63] and attributes of pain (e.g., location, onset, and severity) from clinical notes [12]. Machine learning methods, such as BiLSTM (Bidirectional Long Short-Term Memory) -CRF and SVM (Support Vector Machine), have been employed to extract symptoms of digestive diseases from clinical notes [46], and symptoms and severity of mental illness from hospital discharge summaries and doctor reports [20, 26]. Performance of most of these traditional approaches depends heavily on feature engineering to construct various linguistic features, such as part of speech (POS), word shape, word position in sentence, and local contexts (e.g., words before and after a focal word) [68].

Built on the Transformer architecture [62], BERT [13] is a language model consisting of multiple self-attention layers and millions of weights. BERT is pretrained on massive corpora of books and Wikipedia articles, and can be fine-tuned for a variety of downstream tasks such as document classification, speech recognition, machine translation, and NER [13].

BERT is a generic, open-domain language model with several variants, among which DistilBERT is a smaller and faster version of BERT [55]; RoBERTa builds on BERT using enhanced pretraining strategies [39]; DeBERTa [22] improves the attention and mask decoder in RoBERTa. These BERT-based models automatically construct features by creating context-sensitive word embeddings [3] and generally outperform traditional NER approaches [2, 35, 59] and even ChatGPT [16]. For example, RoBERTa and DeBERTa were featured prominently in many participating systems in the 2023 SemEval NER competition [16]; DistilBERT is reported to perform on-par with medical BERT models using less time and less disk space for medical concept recognition [1].

Compared with GPT-based LLMs, such as LLaMa [61], BLOOM [60], and PaLM [11], which often have billions of parameters, BERT-based models are small. Yet, they can deliver satisfactory performance and are significantly less computationally expensive than other GPT-based LLMs [69].

When used in medical and healthcare applications, generic BERT-based models benefit from fine-tuning on medical documents because most medical terminology is not commonly present in open-domain corpora on which BERT models are trained [31]. A few domain-specific variants, including BioBERT [31], ClinicalBERT [3], and BlueBERT [45], are specifically fine-tuned on medical documents and have shown better performance on biomedical text processing than the generic BERT. Among them, BioBERT, trained using academic papers, performs significantly better than the generic BERT [31], and outperforms other specialized models on biomedical concept recognition (e.g., diseases, chemicals, and proteins) [59].

Although BERT and its variants have shown promising performance, challenges remain when they are used on non-formal, colloquial text. Compared with other medical entities with formal names, symptoms are difficult to recognize in general, and even more so in colloquial writings. Biomedical BERT variants are trained exclusively on formal documents, such as paper abstracts [31, 38], and their performance drops in the presence of errors or noise in the data [4, 25] and they are vulnerable to problems in colloquial text [44]. Furthermore, annotated data sets for symptom recognition are scarce and costly to generate. Therefore, it is desirable to design strategies for effective training using small amounts of labeled informal text data. Unfortunately, little research can be found in the literature addressing the problem of symptom recognition from informal text [40, 41]. Our research addresses this challenge by developing training strategies to fine-tune BERT-based models to effectively recognize concepts and entities in colloquial texts.

## 2.2 Model Performance and Training Strategies

Research has shown that BERT-based models may not perform well when the distribution of data (e.g., terminology used, percentage of sentences with entities) shifts across datasets [17, 28, 52], and that performance of models trained on formal articles usually degrades when applied to non-formal text such as social media posts [44]. The performance drop may be a result of the models' failure to make inferences based on entity contexts [17, 28, 52]. For the NER task, some studies have found model performance to be directly correlated with the overlap of entities between the training and test datasets [17, 28]. For example, an analysis of BERT-based NER in [2] shows that the models mainly memorize the named entities seen in training and rely less on the context in inferring the correct entity type. In other words, these models tend to recall entities seen in the training data, but are less capable of recognizing unseen, out-of-distribution entities in the test data [17, 37].

The ability of a trained model to handle previously unseen data, i.e. model *generalizability*, is an important requirement for machine learning models [18]. Performance and generalizability of BERT-based models depend to a large extent on the training data and strategies rather than on the model architecture alone [14, 37]. Hence, a way to greater generalizability may be through designing more effective training strategies and datasets.

Another possible vulnerability of BERT-based models is sensitivity to "noise" (e.g., spelling errors, typos, and synonyms) that can cause misclassification errors. Experiments reported in [37] show a 35%-62% drop in the F1 score under different *adversarial attacks*, which are intentionally designed noisy test samples. To counter this sensitivity to noise, a type of strategy called *adversarial training* injects training samples perturbed with synonyms and noise into training data to boost the generalizability of models [19]. Adversarial training can better regularize deep learning models than the dropout approach [19], and allows models to adapt to a shifted distribution in the test data improving out-of-distribution generalization [52]. The resulting models tend to be more robust against noise and perturbations in data [10, 19].

A *normalization* strategy, which maps non-formal descriptions of symptoms to formal medical terms [36], can be viewed as a type of adversarial training strategy. For example, the term, *head spinning a little,* can be replaced by the formal term *dizziness* [36]. A deep-learning based entity normalizer was developed in a recent study [41] to convert symptoms and diseases identified from social media messages to UMLS terminology. However, this normalization strategy does not completely address the problem of symptom recognition from colloquial text *per se*, because the informal terms must be identified first before they are mapped to formal medical terms.

Research presented in this paper investigates how different training strategies, which are used to fine-tune BERT and its variants, affect model performance for symptom recognition from colloquial text in organizational settings where computational and storage cost must be considered. Although GPT-based LLMs have shown stunning performance in many tasks such as text generation, coding, and arithmetic reasoning (e.g., [61]), they are much more computationally expensive and a recent comprehensive survey of LLMs shows that they do not perform better than fine-tuned BERT-based models in NER tasks [69]. Notably, several parameter-efficient tuning strategies have been proposed in the literature, such as prefix-tuning [34], prompt tuning [32], low-rank adaptation [24]. Since these strategies require either sophisticated input optimization or injecting low-rank matrices into the layers in the model architecture, they are beyond the scope of this study and may not be an option in organizations with a lack of NLP expertise. In this research, we focus on design choices that can be readily implemented in symptom recognition applications, such as selection of the base model, design of datasets for fine-tuning, the use of perturbation methods.

## 3. Research Design

We employ the design science methodology [23] in our research. Like other design science studies pertaining to machine learning models, our study includes three major phases: data collection and pre-processing, model fine-tuning (training), and performance evaluation. Fulfilling the design science research requirements on *relevance* and scientific *rigor* [23] we collected data from diverse sources, selected widely used BERT-based models in NER research and practice, and designed the training strategies and evaluation methods based on the existing literature. Section 4 will provide details of our datasets and evaluation methods. This section focuses on the design of our model training (i.e., fine-tuning) strategies.

We consider training strategies based on three design choices seen in BERT studies (e.g., [31]): the *base model*, *training corpus*, and *perturbation methods*. The first aspect of the design involves selection of the pretrained base model. Given our focus on extracting symptoms from colloquial texts, we selected six widely adopted pretrained BERT-based models: (1) BERT [13], (2) RoBERTa [39], (3) DeBERTa [22], (4) DistilBERT [55], (5) BioBERT [31], and (6) Bio-Epidemiology-NER [51], which we will refer to as Bio-eNER in the following sections.[2] We chose these models because they have been reported in literature as top performers in the generic NER (e.g., [16]) and biomedical NER tasks (e.g., [1]). Models (1)-(4) are generic, open-domain models, model (5) is a variant of BERT trained specifically for biomedical NLP, and model (6) is a bio-medical NER model built on DistilBERT. We used each model's base version with 12 encoder layers, 768 hidden units at each layer, and 110M weights.

The second aspect of the strategy regards the selection of text corpora with different types of data [13]. We consider three options: (1) colloquial text from social media, (2) formal, academic articles, and (3) a combination of formal and colloquial texts.

The third aspect is related to the adversarial training strategy in which the training data are perturbed to enhance model generalizability [4, 19]. We consider two types of perturbations: *normalization* and *de-normalization*. *Normalization* refers to replacing a colloquial symptom term by a formal medical term with the same POS role (e.g., *difficulty of breathing* → *dyspnea*, *throw up* → *vomit*). This resembles the synonym replacement perturbation [25] or an entity-level adversarial attack [37], by which the entity is altered yet its local context remains unchanged. *De-normalization* refers to another entity-level perturbation: replacing formal symptom names with informal descriptions found in colloquial texts (e.g., *dyspnea* → *difficulty of breathing*).

Based on the three aspects, we designed seven training strategies to fine-tune the six selected models. These strategies are distinguished by their use of different types of training data:

(1) Colloquial text, in which all sentences are non-formal, colloquial narratives written by patients. Since all base models are pretrained using formal texts, this strategy helps "familiarize" them with informal patterns present in colloquial text.
(2) Formal text, in which all sentences are drawn from formal texts. This choice is motivated by the availability of medical texts written in correct, standard English with formal medical terms.
(3) Normalized colloquial text, in which colloquial symptom terms in (1) are replaced by formal terms. This strategy intends to leverage the base model's ability to recognize formal entities, when surrounded by informal context.
(4) De-normalized formal text, in which formal symptoms in (2) are replaced with colloquial terms. This strategy is intended to leverage the base model's ability, if applicable, to recognize the informal symptoms based on the formal contexts.
(5) A mix of formal and colloquial text, where 50% of the sentences are randomly drawn from the colloquial sentences in (1) and the rest from the formal sentences in (2). This strategy mixes the two sets of data with different symptom distributions such as percentages of sentences with symptoms, maximum length of symptom terms, and writing styles (formal vs. colloquial).
(6) Normalized mixed text, in which the informal symptom terms in (5) are normalized.

---

[2] We also evaluated BERTweet, a model trained specifically for processing posts on social media (e.g., Twitter). However, its performance was significantly worse than the performance of the selected models. Due to the paper page limit, we decided not to report its performance metrics in this paper.

(7) De-normalized mixed text, in which formal symptom names in (5) are de-normalized.

## 4. Data and Methods

### 4.1 Data

We collected text data on topics related to the COVID-19 pandemic. We focused on COVID-19 due to its massive impact on society and its wide variety of symptoms. In addition to common flu-like symptoms (e.g., *headache* and *sore throat*), COVID-19 viruses and the vaccines may also affect and cause symptoms in many other body parts and organs [8].

**Training sets.** Corresponding to the seven training strategies, we developed seven training datasets. (1) **CLQ** is a sample of social media messages posted between March and July 2020 on the public forums dedicated to COVID-19 at Patient.info and PatientsLikeMe.com websites. These texts are narratives by patients or their caregivers sharing and describing symptoms, feelings, reactions to treatments, recovery process, and post-traumatic experiences. The writing style is colloquial, and very different from formal writings. Patients often use non-medical terms and figurative expressions; the messages often contain typos, spelling errors, and other noise. Table 1 illustrates three types of challenges for symptom recognitions with examples from the CLQ dataset.

**Table 1.** Examples of colloquial symptom references. Symptoms are highlighted.

| Colloquial Symptom References | Examples from Patient Forum Posts |
|---|---|
| **Non-Medical Terms** | I can't get a deep breath. <br> He started clearing his throat, and had sniffles, then runny nose. <br> I have been having a slight heaviness in the chest area (sometimes). |
| **Figurative Expressions** | Waking up like I'm on fire. <br> Then sharp pains felt like it was flipping from one side of my chest to the other. <br> It feels like I got inside a submarine and submerged a thousand feet under water. |
| **Typos, Abbreviations, Styling Variants** | Mine when [sic] from sore thoat [sic] to breatless [sic] a cough then temp. <br> I am able to breath [sic] normally, but chest or ribs are still paining little bit <br> He has light coughs. |

A message was collected if it contained at least one sentence mentioning a symptom. The resulting CLQ dataset contains 1,384 sentences, 33% of which mention at least one symptom. The longest symptom has 16 tokens (e.g., words and punctuation marks), and 49% of the symptoms are one-word long. (2) **CORD** is a dataset with a sample size comparable to that of CLQ. This sample contains 1,472 sentences that we extracted between May 1st and 10th 2020 from the COVID-19 Open Research Dataset, which contains over 500,000 scholarly articles related to COVID-19 [64]. Most sentences (92%) in this sample contain at least one symptom. The longest symptom has 9 words, though 80% of the symptoms are one-word long. (3) **CLQ-N** is the CLQ dataset with all colloquial symptoms manually normalized to their formal terms. (4) **CORD-D** is the CORD dataset with all formal symptom names de-normalized to informal terms from CLQ or the Internet. (5) **MIX** contains 1,426 sentences, half of which randomly drawn from the CLQ set, and the rest from the CORD set. (6) **MIX-N** is the normalized version of MIX. (7) **MIX-D** is the de-normalized version of MIX.

**Test sets**. Four test datasets were collected from different sources: (1) **CORD-Test** contains 275 sentences randomly extracted from the COVID-19 Open Research Dataset. There is no overlap between CORD and CORD-Test. (2) **CLQ-Test** contains 131 sentences posted between August 2020 and April 2022 on Patient.info and PatientsLikeMe.com. (3) **MedHelp** contains 2,820 sentences posted between March 2020 and August 2022 in the COVID-19 forum on MedHelp.com, a social media platform for patients to share experiences. Only 18% of the sentences have symptoms. (4) **iCliniq** is a sample of 6,295 sentences posted between March 2020 and August 2022 in the COVID-19 groups on iCliniq.com, an online platform where patients can ask questions and get answers from registered physicians. Approximately 26% of the sentences have symptoms. Table 2 presents the statistics of the seven training and four test datasets and the distributions of symptom entities.

**Table 2.** Statistics of training and test datasets.

| Datasets | | # Sentences | # Symptoms | # Distinct Symptoms | % Sentences with Symptoms | Max Symptom Length | % One-word Symptoms |
|---|---|---|---|---|---|---|---|
| Training Sets | CLQ | 1,384 | 924 | 466 | 33% | 16 | 49% |
| | CORD | 1,472 | 2,910 | 322 | 92% | 9 | 80% |
| | CLQ-N | 1,384 | 913 | 230 | 33% | 9 | 64% |
| | CORD-D | 1,472 | 2,905 | 432 | 92% | 14 | 44% |
| | MIX | 1,426 | 1,878 | 445 | 62% | 16 | 73% |
| | MIX-N | 1,426 | 1,863 | 305 | 62% | 9 | 77% |
| | MIX-D | 1,426 | 1,863 | 481 | 62% | 16 | 43% |
| Test Sets | CORD-Test | 275 | 545 | 179 | 89% | 7 | 73% |
| | CLQ-Test | 131 | 128 | 78 | 44% | 8 | 47% |
| | MedHelp | 2,820 | 836 | 369 | 18% | 12 | 55% |
| | iCliniq | 6,295 | 2,664 | 836 | 26% | 16 | 54% |

Examining these data sets we found that symptoms in colloquial data from patient forums include standard medical terms like *fever*, *cough, headache*, among most commonly occurring, but also contain a broad variety of informal symptom descriptions, using long verb-phrases and figurative language (see Table 1 for examples). These long terms are usually unique yet might contain body parts or words found in formal symptom terms (e.g., *taste*, *smell*). In contrast, symptoms in formal texts are described using mostly single nouns (e.g., *dyspnea*) and short noun phrases (e.g., *chest pain*). In all colloquial and de-normalized data sets used in training (CLQ, CORD-D, MIX-D) the percentage of one-word symptoms is under 49%, while at least 73% of symptoms in the formal and mixed data sets (CORD, MIX) have only a single word.

We employed the IOB (Inside-Outside-Beginning) tagging scheme [50] to label the datasets. Each word in a sentence was manually tagged as B-SYM (beginning of a symptom), I-SYM (inside of a symptom), or O (outside of a symptom). For example, the sentence, "*I have a bad headache and cough a lot*" is tagged as [O, O, O, B-SYM, I-SYM, O, B-SYM, I-SYM, I-SYM]. Since we focused on symptoms, all other entities, such as persons, organizations, diseases, and treatments, were labelled as O. The CORD datasets were manually labelled based on a list of 329 formal names for COVID-19 related symptoms found in the literature [63]. Each of the colloquial datasets was independently tagged by two research assistants. The average inter-coder agreement was 98%. Disagreements were resolved by discussions.

*4.2 Methods*

We employ *precision*, *recall*, and *F1 score*, which are standard performance metrics used in information retrieval and NER studies, to evaluate model performance in this research:

$Precision = \frac{\text{\# Correctly Recognized Symptoms}}{\text{Total \# Recognized Symptoms}} = \frac{TP}{TP+FP}$,

$Recall = \frac{\text{\# Correctly Recognized Symptoms}}{\text{Total \# True Symptoms}} = \frac{TP}{TP+FN}$,

$F1\ score = \frac{2 \times Precision \times Recall}{Precision+Rec}$,

where TP, FP, and FN are the number of True Positives, False Positives, and False Negatives, respectively.

As commonly done in NER literature [66], we compute precision, recall and F1 using two types of matches between the true label sequence and the predicted one: (1) *Exact match*: a recognized symptom term is considered correct only if the sequence of predicted labels in the term matches the sequence of true labels and its boundary exactly [54]. (2) *Partial (*a. k. a. *relaxed) match*: a symptom term is considered recognized if they overlap on at least one word.

We evaluate the performance of the *six* base models fine-tuned using the *seven* training sets on the *four* test datasets. To establish a baseline, we also include in the comparison the SignSym system, a state-of-the-art lexicon-based NER model specifically designed to recognize COVID symptoms from text [63]. The selected models were implemented and evaluated on Google Colab, a cloud-based Python application development environment, using T4 GPUs. Each model was fine-tuned for 4-6 epochs using the AdamW optimizer and a learning rate of 3E-5.

## 5. Analysis and Results

In this section, we report model performance metrics, summarize the effects of different design choices on performance, and examine the classification results with an in-depth analysis.

### 5.1 Model Performance

The six selected models were fine-tuned on the seven training sets, resulting in 42 fine-tuned models. Table 3 reports F1 scores based on exact matches, as well the partial F1 (pF1) from our experiments testing these models on unseen formal and colloquial data. The SignSym column reports the F1 scores from running the SignSym system [63] on the same test sets. The complete set of performance metric values based on exact match including precision and recall appears in the Appendix.

Training sets in Table 3 are arranged in order of increasing share of colloquial content: from CLQ (purely colloquial), MIX (a mix of colloquial and formal data), to CORD (purely formal). Among the four *test* sets, CORD-Test is completely formal, and the other three are colloquial. For each test set in Table 3 and the following tables, the best score in each row (i.e., training set) is highlighted in bold face, and the second-best in italics. The best score on the test set across all model types and training sets is underlined and boldfaced.

**Table 3.** Results of model performance testing on unseen data: (1) BERT, (2) RoBERTa, (3) DeBERTa, (4) DistilBERT, (5) BioBERT, and (6) Bio-eNER.

| Test Set | SignSym | Training Set | F1 | | | | | | pF1 | | | | | |
|---|---|---|---|---|---|---|---|---|---|---|---|---|---|---|
| | | | (1) | (2) | (3) | (4) | (5) | (6) | (1) | (2) | (3) | (4) | (5) | (6) |
| **CORD-Test** | 0.63 | CLQ | 0.59 | *0.77* | 0.75 | 0.62 | **0.78** | 0.74 | 0.79 | 0.84 | 0.83 | 0.82 | **0.86** | *0.85* |
| | | CLQ-N | 0.65 | **0.79** | **0.79** | 0.67 | **0.79** | *0.77* | 0.82 | **0.88** | *0.86* | 0.84 | *0.86* | *0.86* |
| | | MIX-D | 0.72 | **0.78** | *07.7* | 0.76 | **0.78** | 0.76 | 0.81 | **0.85** | *0.84* | *0.84* | **0.85** | **0.85** |
| | | **MIX** | 0.75 | 0.79 | 0.78 | 0.75 | <u>**0.81**</u> | *0.80* | 0.85 | **0.89** | 0.87 | *0.88* | **0.89** | **0.89** |
| | | **MIX-N** | 0.74 | *0.80* | 0.79 | 0.75 | <u>**0.81**</u> | *0.80* | 0.85 | *0.88* | *0.88* | *0.88* | *0.88* | **<u>0.90</u>** |
| | | CORD-D | 0.74 | **0.80** | 0.78 | 0.77 | 0.78 | *0.79* | 0.83 | **0.86** | *0.85* | 0.84 | 0.84 | **0.86** |
| | | CORD | 0.76 | **0.80** | *0.79* | 0.77 | *0.79* | **0.80** | 0.86 | **0.89** | *0.88* | **0.89** | *0.88* | **0.89** |
| **CLQ-Test** | 0.57 | **CLQ** | 0.56 | <u>**0.82**</u> | *0.81* | 0.63 | 0.74 | 0.72 | 0.87 | **0.93** | *0.92* | 0.88 | 0.87 | 0.89 |
| | | CLQ-N | 0.55 | 0.75 | *0.77* | 0.60 | **0.78** | 0.73 | 0.85 | **0.91** | **0.91** | 0.88 | 0.88 | *0.90* |
| | | MIX-D | 0.64 | 0.75 | *0.77* | 0.66 | **0.78** | 0.72 | 0.86 | **0.92** | *0.90* | 0.88 | 0.88 | 0.88 |
| | | **MIX** | 0.63 | *0.75* | **0.76** | 0.65 | **0.76** | 0.74 | 0.88 | <u>**0.94**</u> | *0.91* | 0.87 | 0.90 | 0.90 |

|  |  |  |  |  |  |  |  |  |  |  |  |  |  |
|---|---|---|---|---|---|---|---|---|---|---|---|---|---|
|  |  | MIX-N | 0.61 | *0.75* | *0.75* | 0.69 | 0.73 | **0.76** | 0.86 | **0.91** | **0.91** | 0.88 | 0.89 | *0.90* |
|  |  | CORD-D | 0.58 | **0.72** | *0.71* | 0.55 | *0.71* | 0.62 | 0.79 | **0.87** | **0.87** | 0.78 | *0.85* | 0.77 |
|  |  | CORD | 0.66 | 0.71 | *0.72* | 0.64 | **0.74** | *0.72* | 0.87 | **0.91** | **0.91** | 0.85 | *0.87* | **0.91** |
| **MedHelp** | 0.49 | **CLQ** | 0.52 | *0.69* | **0.71** | 0.52 | 0.62 | 0.59 | 0.74 | *0.82* | **0.83** | 0.78 | 0.77 | 0.79 |
|  |  | CLQ-N | 0.53 | *0.68* | **0.70** | 0.55 | 0.63 | 0.64 | 0.77 | **0.84** | *0.83* | 0.81 | 0.78 | 0.82 |
|  |  | MIX-D | 0.58 | *0.67* | **0.68** | 0.59 | 0.62 | 0.61 | 0.76 | *0.83* | **0.84** | 0.79 | 0.79 | 0.78 |
|  |  | **MIX** | 0.56 | *0.67* | **0.69** | 0.58 | 0.63 | 0.60 | 0.79 | **0.85** | *0.84* | 0.81 | 0.82 | 0.79 |
|  |  | MIX-N | 0.55 | **0.68** | 0.68 | 0.59 | *0.62* | 0.64 | 0.78 | **0.85** | *0.84* | 0.83 | 0.81 | 0.82 |
|  |  | CORD-D | 0.46 | **0.59** | 0.59 | 0.44 | *0.52* | 0.45 | 0.71 | *0.74* | **0.75** | 0.62 | 0.75 | 0.60 |
|  |  | CORD | 0.57 | **0.64** | 0.64 | 0.57 | 0.62 | *0.63* | 0.79 | *0.80* | **0.81** | 0.81 | *0.80* | **0.81** |
| **iCliniq** | 0.53 | **CLQ** | 0.51 | *0.68* | **0.70** | 0.54 | 0.64 | 0.59 | 0.76 | *0.80* | **0.82** | 0.78 | 0.76 | 0.78 |
|  |  | CLQ-N | 0.52 | *0.65* | **0.66** | 0.56 | 0.64 | 0.62 | 0.79 | *0.81* | **0.82** | 0.80 | 0.79 | **0.82** |
|  |  | MIX-D | 0.60 | *0.67* | **0.68** | 0.60 | 0.65 | 0.65 | 0.77 | **0.83** | **0.83** | 0.79 | 0.77 | *0.82* |
|  |  | **MIX** | 0.57 | *0.65* | **0.66** | 0.58 | 0.64 | 0.61 | 0.79 | *0.83* | **0.84** | 0.81 | 0.80 | 0.81 |
|  |  | **MIX-N** | 0.55 | **0.65** | **0.65** | 0.59 | 0.63 | *0.64* | 0.78 | **0.84** | **0.84** | *0.81* | 0.80 | **0.84** |
|  |  | CORD-D | 0.54 | *0.62* | **0.63** | 0.54 | 0.60 | 0.57 | 0.62 | *0.78* | **0.79** | 0.73 | 0.68 | 0.73 |
|  |  | CORD | 0.59 | *0.62* | 0.61 | 0.58 | *0.62* | **0.63** | 0.76 | *0.82* | 0.81 | 0.80 | 0.78 | **0.83** |

Table 3 shows that the median scores of all six fine-tuned BERT-based models on all four test sets are higher than the scores of SignSym. In the following, we highlight results from our experiments, focusing on the effects of different training strategies on model performance. We refer to models trained on CORD data as formal models, those trained on CLQ data as colloquial models, and MIX, as mixed models, for brevity. We also use *entity*, *term*, and *symptom* interchangeably henceforth.

### 5.2 Effects of Design Choices for Training Strategies

#### 5.2.1 Overall performance across base-model types

Observing results in Table 3, we conclude that when tested on formal data (CORD-Test), the domain-specific models BioBERT and Bio-eNER produce the best F1 and pF1 scores, respectively. It is noteworthy that these best scores on the formal test set result from fine-tuning the biomedical models on the *mixed* data (MIX and MIX-N) instead of the *formal* data (CORD, CORD-D). On the colloquial CLQ-Test, RoBERTa fine-tuned on CLQ data has the highest F1 score, while on MedHelp and iCliniq, the best F1 scores are produced by DeBERTa trained on CLQ. RoBERTa trained on MIX or MIX-N produces the highest pF1 scores on all colloquial tests, although on the iCliniq test, it is tied with DeBERTa and Bio-eNER trained on one or both of MIX or MIX-N datasets. The two models showing best scores on colloquial tests, DeBERTa and RoBERTa, have similar F1 and pF1 scores over most tests.

#### 5.2.2 Generic Models vs. Domain-specific Models

Columns (5) and (6) in Table 3 correspond to BioBERT and Bio-eNER, which are the domain-specific, biomedical variants of BERT and DistilBERT, presented in columns (1) and (4), respectively. Comparing F1 scores from the two latter base models and their specialized variant in each row in Table 3, shows that on colloquial and formal tests the domain-specific models outperform their generic counterparts fine-tuned using the same training data.

#### 5.2.3 Colloquial vs. Formal vs. Mixed Training Data

To get a summary view of models fine-tuned on different training sets across all three performance metrics, we use a performance scoring approach. The scoring counts the number of times a model fine-tuned on a specific training set produces one of the top-two metrics (exact or partial precision, recall, or F1) across all models of its type (see Appendix Table 1). We call these counts *Score* and *pScore* and present them for the four top-scoring mode types in Table 4. For example, the *Score* for the BioBERT-based CLQ model (in the first row of Table 4) is 5, because according to Appendix Table 1, in all three colloquial tests, its precision (P) is never in top-two, recall (R) score is among the top-two 3 times, and its F1 score 2 times (on MedHelp and iCliniq). This model's *pScore* is 3, comprised of 0 top-two pP and pF1 values and 3 top-two pR values. The Total Score is the sum of Score and pScore.

**Table 4.** Score (P, R, F1), pScore (pP, pR, pF1) and their sum (Tot) for models assessed over the colloquial and formal test data. Bold face indicates the highest value in each column per test.

| Test data | Training Set | BioBERT | | | DeBERTa | | | RoBERTa | | | Bio-eNER | | |
|---|---|---|---|---|---|---|---|---|---|---|---|---|---|
| | | Score | pScore | Tot | Score | pScore | Tot | Score | pScore | Tot | Score | pScore | Tot |
| CLQ-Test + MedHelp +iCliniq | CLQ | 5(0,3,2) | 3(0,3,0) | 8 | **9(3,3,3)** | 5(0,3,2) | 14 | 8(2,3,3) | 4(0,3,1) | 12 | 1(0,1,0) | 0(0,0,0) | 1 |
| | CLQ-N | 4(0,1,3) | 2(0,1,1) | 6 | 5(3,0,2) | 4(2,0,2) | 9 | 4(2,0,2) | 3(2,0,1) | 7 | 4(2,1,1) | 4(2,0,2) | 8 |
| | MIX-D | **6(0,3,3)** | 3(0,3,0) | 9 | 6(3,1,2) | 4(2,0,2) | 10 | 3(1,0,2) | 2(1,0,1) | 5 | 5(1,3,1) | 2(0,2,0) | 7 |
| | MIX | 5(2,0,3) | 6(2,1,3) | 11 | 3(1,2,0) | 7(1,3,3) | 10 | 4(0,3,1) | 7(1,3,3) | *11* | 3(0,2,1) | 4(0,3,1) | 7 |
| | MIX-N | 3(2,0,1) | 6(3,0,3) | 9 | 2(2,0,0) | 6(3,0,3) | 8 | 3(1,0,2) | 5(3,0,2) | 8 | **6(3,0,3)** | 6(3,0,3) | 12 |
| | CORD-D | 0(0,0,0) | 1(0,1,0) | 1 | 0(0,0,0) | 2(0,2,0) | 2 | 1(0,1,0) | 1(0,1,0) | 2 | 3(0,3,0) | 3(0,3,0) | 6 |
| | CORD | 3(2,0,1) | 3(3,0,0) | 6 | 0(0,0,0) | 2(1,0,1) | 2 | 0(0,0,0) | 1(1,0,0) | 1 | 3(2,0,1) | 6(3,0,3) | 9 |
| CORD-Test | CLQ | 1(0,1,0) | 1(0,1,0) | 2 | 0(0,0,0) | 0(0,0,0) | 0 | 1(0,1,0) | 1(0,1,0) | 2 | 0(0,0,0) | 1(0,1,0) | 1 |
| | CLQ-N | 2(0,1,1) | 1(0,1,0) | 3 | 2(0,1,1) | 1(0,1,0) | 3 | 3(1,1,1) | 2(0,1,1) | 5 | 1(0,1,0) | 0(0,0,0) | 1 |
| | MIX-D | 0(0,0,0) | 0(0,0,0) | 0 | 1(1,0,0) | 0(0,0,0) | 1 | 1(1,0,0) | 0(0,0,0) | 1 | 0(0,0,0) | 0(0,0,0) | 0 |
| | MIX | 3(1,1,1) | 2(1,0,1) | 5 | 2(0,1,1) | 3(1,1,1) | 5 | 2(0,1,1) | 3(1,1,1) | 5 | 2(0,1,1) | 2(0,1,1) | *4* |
| | MIX-N | 3(1,1,1) | 2(1,0,1) | 5 | 2(0,1,1) | 3(1,1,1) | 5 | 3(1,1,1) | 3(1,1,1) | 6 | 3(1,1,1) | 2(1,0,1) | 5 |
| | CORD-D | 0(0,0,0) | 0(0,0,0) | 0 | 2(1,0,1) | 1(1,0,0) | 3 | 3(1,1,1) | 0(0,0,0) | 3 | 2(1,0,1) | 0(0,0,0) | 2 |
| | CORD | 1(0,0,1) | 2(1,0,1) | 3 | 2(0,1,1) | 3(1,1,1) | 5 | 3(1,1,1) | 3(1,1,1) | 6 | 3(1,1,1) | 2(1,0,1) | 5 |

According to Table 4, for the listed base model types, the training sets that result in the simultaneous best or second-best total score for all models, as judged by the sum of Score and pScore (column Tot) for colloquial and formal tests, are either MIX or MIX-N. Specifically, BioBERT achieves two of its top scores when fine-tuned on MIX, MIX-N or MIX-D (colloquial test), and MIX or MIX-N (formal test). DeBERTa's top two models for colloquial tests are CLQ and MIX or MIX-D, and for the formal test – MIX, MIX-N or CORD, and so forth for RoBERTa and Bio-eNER. This means that if a model needs to be simultaneously optimized for formal and colloquial data, fine-tuning on either MIX or MIX-N should work best.

### 5.2.4 Normalization and De-normalization

There are notable differences in the effects of normalization and de-normalization perturbations on model precision and recall. In our results (see Appendix Table 1), normalization of CLQ and MIX data resulted in an increase in precision for CLQ-N and MIX-N models in 72% and 78% of results, respectively, on colloquial tests (3 test datasets, 6 model types). De-normalizing entities in formal training data (CORD) improves recall on colloquial tests in 100% of cases, while effect of de-normalization of mixed data set MIX is not as strong, increasing recall values in 39% of cases. Similar effects of normalization and de-normalization are observed on the partial measures (pP and pR).

### 5.3 In-depth Analysis

This section presents an in-depth analysis of model performance and types of classification errors. As extraction of new, rare, and long colloquial terms is an important aspect of symptom recognition from colloquial data [40], we compare performance of models for recognizing terms of different lengths and take a closer look at the model outputs for terms that are recognized correctly (TP: True Positive), incorrectly (FP: False Positive), and those terms that are missed (FN: False Negative).

*5.3.1 True Positives (TP)*

On both colloquial and formal tests, the total number of symptoms correctly identified by CLQ-trained models is higher than those of MIX-trained, which, in turn, is higher than terms found by the CORD-trained models, with a single exception: on formal test DeBERTa MIX-trained model finds 2 fewer terms than the CORD-trained one. We do not have space to present these counts, and instead include a summary in Table 5, where results are aggregated across all models trained on a specific dataset. Table 5 presents ranges of percentages of true symptoms of specific length correctly recognized by models trained on each of the training datasets by exact match. For example, TP-1 indicates True Positive of single-word terms, while TP-4+ of length four or longer.

**Table 5.** Ranges of proportions of correctly identified symptoms by length (percent of symptoms of specific length correctly recognized with exact matching) in colloquial and formal tests, as identified by all models trained on the specific test. TP-N indicates True Positive of length N.

| Training Set | Colloquial Test Data | | | | Formal Test Data | | | |
| --- | --- | --- | --- | --- | --- | --- | --- | --- |
| | TP-1 | TP-2 | TP-3 | TP-4+ | TP-1 | TP-2 | TP-3 | TP-4+ |
| CLQ | 89-96% | 67-78% | 54-73% | 25-50% | 80-95% | 73-84% | 43-71% | 12-59% |
| CLQ-N | 90-94% | 62-72% | 38-54% | 13-26% | 83-95% | 71-83% | 39-64% | 6-59% |
| MIX-D | 90-94% | 69-75% | 50-60% | 15-31% | 80-89% | 69-77% | 39-46% | 6-18% |
| MIX | 92-95% | 56-68% | 42-62% | 21-37% | 88-95% | 58-67% | 36-50% | 6-18% |
| MIX-N | 92-94% | 51-66% | 30-49% | 10-20% | 88-93% | 55-64% | 36-46% | 6-18% |
| CORD-D | 93-95% | 72-75% | 47-59% | 13-25% | 84-90% | 68-74% | 39-50% | 12-18% |
| CORD | 89-94% | 46-60% | 29-40% | 5-16% | 88-93% | 48-62% | 36-43% | 0-12% |

Table 5 shows that CLQ-tuned models recognize more of the longest (TP-4+) symptoms than others, while the CORD models recognize the fewest such symptoms. Fine-tuning on CORD leads to fewer multi-word terms than on its de-normalized version CORD-D. These results evidence that colloquial data in pure form, or the use of colloquial terms in de-normalization of formal training data may improve a model's ability to recognize multi-word symptoms in both formal and colloquial texts.

Table 6 presents the longest symptom terms recognized by each of the models in the combined colloquial data based on exact matches. We found that a majority of the longest terms recognized by CORD are related to loss of smell and taste across all model types, and that other training sets yielded a much greater variety of correctly recognized long symptoms than CORD.

It is evident from the analyses that CLQ has a greater capacity for capturing long and rare symptoms, compared to MIX and CORD, and MIX has the same advantage, though to a lesser degree, over CORD. It is possible that the prevalence of one-word symptoms in the formal data set (CORD) predisposes models trained on it towards predicting single word terms. Presence of longer, colloquial terms in MIX set mitigates this limitation of formal training data sets and improves the models' ability to detect terms of a greater variety and greater length. The model trained on the purely colloquial set (CLQ) misses very few common short terms detected by the MIX and CORD. This is likely due to the most common symptoms from the formal data, such as *fever, sore throat, fatigue*, and others being very prominently represented in the CLQ patient narrative data.

**Table 6.** Longest symptoms (and sizes) from the combined colloquial test data correctly recognized by (1) BERT, (2) RoBERTa, (3) DeBERTa, (4) DistilBERT, (5) BioBERT, and (6) Bio-eNER.

|  | (1) | (2) | (3) | (4) | (5) | (6) |
|---|---|---|---|---|---|---|
| CLQ | $9^1$ | $11^2$ | $10^3$ | 8 | $13^4$ | 8 |
| CLQ-N | 6 | 7 | 7 | 6 | 7 | 6 |
| MIX-D | 9 | 10 | 10 | 7 | 13 | 8 |
| MIX | 9 | 9 | 10 | 8 | 13 | 9 |
| MIX-N | 6 | 7 | 7 | 6 | 6 | 6 |
| CORD-D | 7 | 7 | 8 | 7 | 7 | 8 |
| CORD | 5 | 6 | 6 | 5 | 6 | 6 |

1: "tingling and burning sensation in my arms and legs"
2: "can not taste, smell or sense any change in temperature"
3: "left ear hurts a lot when I swallow my saliva"
4: "pain, especially in his hand, finger joints, and feet joints"

*5.3.2 Error Analysis: False Positives (FP)*

As an illustration, we report findings from examining the BioBERT models. We find that FP terms are predominantly one-word long, and de-normalization increases the FP counts, while normalization reduces FP counts in both CLQ-N and MIX-N compared to CLQ and MIX.

We observe that for both MedHelp and iCliniq, the number of FP terms produced by colloquial CLQ is greater than that of MIX, which is greater than CORD's. For both test sets, the lowest FP counts correspond to the normalized mixed model (MIX-N) with the formal CORD being the close second.

The effects from normalization, de-normalization, and the relationship between the results from formal, mixed, and colloquial models suggest that exposure to a broader vocabulary of symptoms, as well as a greater average symptom length in training set, increases the number of False Positives.

Comparing the FP terms of CLQ and CLQ-N on MedHelp, we found that the CLQ model includes more FPs that refer to body locations (e.g., *in the gut, from neck down*), more qualifiers (e.g., *moist, terrible*), and verb phrases (e.g., *felt like I had been, couldn't walk*).

Among the 78 FP terms incorrectly identified by *all* BioBERT models in the MedHelp test, a large proportion are single-word names of diseases and conditions (e.g., *cancer, pneumonia*) often used to provide the background health information, sometimes in the context similar to that of a symptom description (e.g., *He has cancer*). Some common COVID symptom words, such as *cough, fever,* and *pain*, are sometimes misidentified as symptoms when referring to something else, as in the sentence, *"A single cough releases about 3,000 droplets."*

*5.3.3 Error analysis: False Negatives (FN)*

When tested on colloquial data, the CORD-trained models consistently produced the highest number of False Negatives (FN) for each model type by the partial matching, which indicates the number of true symptoms completely missed by the model. Normalization of both CLQ and MIX data increases the FN counts, while de-normalization of the CORD data reduces it, which is consistent with the impact of these treatments on precision and recall noted earlier.

Comparing the FN output of CORD and CORD-D trained BioBERT to illustrate the effect of de-normalization of formal data on the variety of recognized symptoms, CORD model missed some colloquial expressions for common symptoms correctly identified by CORD-D, for example *"Things smell very weird , off , faint or not at all", mild difficulty taking deep breaths, chest again felt a little tight,* and even the more formal term, *high blood pressure.* CORD-D identified some more unusual short symptoms missed by CORD, such as *brain fog*, *cold hands,* and *twinges.*

## 6. Discussion

Our research is motivated by challenges of entity recognition from colloquial text. More specifically, our study focuses on symptom recognition from patient-written narratives using BERT and its variants. We selected BERT-based variants that have shown outstanding performance in generic and biomedical NER tasks [16, 31, 38, 42, 59] and require significantly less computational resources than GPT-based LLMs [71].

We tackle the challenges by employing several model training strategies with different design choices regarding the selection of the base model, types of training data, and entity perturbation methods. Using design science methodology, we conducted a series of experiments comparing performance of six BERT-based NER models fine-tuned on COVID-19 related training datasets, which were compiled using different strategies. In the following we summarize our findings and propose design guidelines for training strategies appropriate for different goals and conditions. These design choices and guidelines can be easily implemented and integrated in applications for clinical diagnosis or public health monitoring, for symptom recognition from colloquial text to support decision-making in healthcare and medical contexts.

### *6.1 Summary of Findings and Design Guidelines*

**Base Models:** (1) Fine-tuned on colloquial data, RoBERTa and DeBERTa tend to outperform all other models, including the domain-specific BioBERT and Bio-eNER, achieving best performance in F1 score when tested on unseen colloquial data. (2) BioBERT and Bio-eNER achieve their best performance on formal text when fine-tuned on mixed data, which combines colloquial and formal texts. (3) All BERT-based models outperform the SignSym system on formal and on colloquial texts.

**Training Data (Colloquial, Formal, Mixed):** (1) Models fine-tuned using only formal text tend to produce higher precision (except for DeBERTa and RoBERTa) yet lower recall, as they capture most single-word symptoms, and fail to recognize long, rare, and unusual ones. (2) Models fine-tuned on colloquial text best capture long and rare colloquial symptoms and tend to produce higher recall. (3) Models trained on a mix of formal and colloquial data achieve best balance between precision and recall when tested on both formal and colloquial data. Notably, when fine-tuned on mixed data, the domain-specific models achieve higher F1 scores even on unseen formal test data.

**Perturbations (Normalization, De-normalization):** (1) Normalization helps improve precision but reduces recall on colloquial tests. Models trained on normalized data are less able to identify long and unusual terms. (2) De-normalization of formal training data leads to improved recall but reduced precision on colloquial tests. Models fine-tuned on de-normalized formal data can identify more longer colloquial terms than models trained on purely formal text.

Overall, our experiments have shown that using a combination of colloquial and formal data to fine-tune a model improves a model's capabilities of recognizing colloquial symptom phrases, without sacrificing its ability to find commonly used formal terms. Our in-depth analysis of correct classification and misclassification provides strong evidence that the presence of colloquial terms in the training data caused the models to identify more rare, multi-word terms.

Based on these findings, we propose the following design guidelines for developing effective training strategies for symptom recognition from formal or colloquial text. A designer can use these guidelines to choose different strategies based on the goal of the task, performance priority, and availability of labeled training data.

(1) **Task Goal**. Depending on the availability of training data and whether the goal is to recognize symptoms from *colloquial text*, *formal text*, or *both*, different models can be selected.
   a. If the goal is symptom recognition *from colloquial text and labeled colloquial data are available* for fine-tuning, two generic variants of BERT, namely RoBERTa and DeBERTa, are preferable to domain-specific BioBERT or Bio-eNER. Although all six models are pre-trained on formal text, RoBERTa and DeBERTa are more capable of adjusting to non-formal symptom terms and contexts in colloquial text.
   b. If the goal is to recognize symptoms *from formal text*, the domain specific models (i.e., BioBERT or Bio-eNER) fine-tuned on mixed data should be used. These two models are pretrained on biomedical literature and tend to outperform other four generic BERT-based models in symptom recognition from formal texts.
   c. If the goal is to achieve high performance on both *formal and colloquial data*, fine-tuning RoBERTa, DeBERTa, or BioBERT on mixed data tends to deliver comparable best performance.

(2) **Performance Priority**. Depending on the performance priority, whether it be *recall*, *precision*, or both (*F1*), different training data should be used.
   a. If the priority is *recall*, that is, to extract as many entities as possible including rare and long terms, colloquial data should be used to fine-tune the selected base model. Models trained on colloquial data produce higher recall and more multi-word, unusual symptoms than models trained on formal data. However, precision may be relatively low.
   b. If the priority is to ensure *precision*, formal text can be used when labeled colloquial data are not available. However, although they tend to be more precise, models trained on formal data have lower recall, and may miss a large number of long, unusual terms.
   c. If the priority is *balanced recall and precision* (F1), a mixed training set that includes an equal share of sentences from colloquial and formal data achieves a "sweet-spot" between precision and recall, producing best F1 scores on unseen colloquial and formal texts.

(3) **Availability of Labeled Data**. If only formal or only colloquial data are available, *perturbation methods* of normalization or de-normalization can be used to affect precision or recall. Normalization tends to reduce recall and increase precision. De-normalization of formal training data has the opposite effect (improving recall, but reducing precision), leading to a lower F1 score. On the other hand, it helps capture a wider variety of symptoms.

## 6.2 Implications for Research

Our research makes several contributions to the literature: First, this study focuses on an understudied NER task of recognizing entities from colloquial text. Although great advances have been made in recent years, NER from colloquial text has remained a challenge. We seek to tackle the problem of recognizing symptoms, which are more difficult to identify than other medical concepts such as disease and medicine names. By fine-tuning six BERT-based models, we show that the models performed differently but all outperformed a state-of-the-art tool (SignSym) specialized for symptom recognition in this domain.

Second, our experiments provide empirical evidence and findings regarding how different design choices and aspects of training strategies affect model performance, adding new insights to research on performance and generalizability of the BERT family of models. Moreover, our in-depth analyses of the correct predictions and errors produced by the models, based both on exact and partial matches, identify situations where models perform in different dimensions (precision and recall). The analysis of the impacts of the normalization and de-normalization methods provides insights into how these models utilize entity and context information when making inferences in NER tasks.

Third, we designed several training strategies, combining three design choices regarding the base model, training data, and perturbation methods, and proposed a set of design guidelines for making these design choices. Medical concepts and entities extracted can be categorized, aggregated, and analyzed to support various data-based decision-making in the healthcare and medical contexts. For example, detection of rare symptoms unique to a certain disease may allow physicians to develop effective treatment plans. In this sense, our research joins other design science studies in the Information Systems (IS) literature to help develop better healthcare decision-support systems [27, 30, 33, 65].

Through the lens of the design science knowledge framework [21, 23], this research makes a Level-2 contribution to the literature: knowledge as operational principles. Based on our empirical study of the performance of different models, we contribute both *prescriptive knowledge*, in the form of model training strategies and design guidelines, as well as *descriptive knowledge* in the form of patterns of model behaviors observed when trained using these strategies.

It is worth stressing the *relevance* of this research, as well as the value of BERT-based models, especially when GPT-based LLMs seem to attract all the attention. Many believe that there has been a paradigm shift away from the pretrain-and-fine-tune one in NLP research and practice [69]. New LLMs have emerged rapidly and continued to excel in many AI tasks (e.g., text-to-image generation). With zero- or few-shot training and in-context learning, these LLMs can understand prompts in natural languages and generate coherent responses. However, it has been shown that some traditional information retrieval tasks, including NER, remain a challenge for LLMs [47], while the pre-trained and fine-tuned models still are the top choices [69]. Therefore, we believe that the BERT-based models will continue to be relevant and useful for NER tasks due to their reliable performance and other practical considerations, which will be discussed next.

### *6.3 Implication for Practice*

As more and more communication takes place in social media, virtual meetings and visits, instant messages, and live streaming sessions, the information that such communication contains can be extracted to support decision-making in different domains (e.g., to improve customer relationships or learn market trends). Against this backdrop, this study has important implications for the practice of entity recognition from colloquial text in general, and symptom recognition in particular.

First, our study provides empirical evidence on the impacts of different training design choices on model performance. Experimental results presented here show that symptom recognition in colloquial text is more difficult than in formal text, and that addition of labeled colloquial data to the dataset used to fine-tune BERT and its variants visibly improves the model's performance, achieving greater recognition of longer, colloquial terms. Indeed, the presence of a relatively small number of colloquial terms in training data has a noticeable effect on model's performance. Moreover, as it is often desirable to achieve better precision and/or recall, our results suggest that these performance aspects can be improved through appropriate levels of normalization and de-normalization of training data. In addition, we provide performance metrics measured based on partial match, which may be more practically meaningful for applications in which both exact and inexact term detection is of value.

Second, depending on the task goal, performance priorities, and data availability, our design guidelines can help designers choose the appropriate base model, training datasets, and perturbation methods. Given the wide accessibility of BERT-based models and their relatively small sizes, these design choices can be easily adopted and implemented in healthcare organizations, which do not necessarily have extensive computational resources and NLP application development expertise that GPT-based LLMs often require. Organizations can choose their appropriate models, training strategies, and datasets based on their available resources and budgets. The fine-tuned models can be integrated into various clinical and EHR systems, as well as public health monitoring and reporting systems, to extract symptoms and other medical concepts. More importantly, in the medical and healthcare domain, many patient-related decisions are critical. In particular, as GPT-based, generative LLMs sometimes suffer from various problems such as hallucination [71], healthcare professionals are extremely concerned with the ethical implications of adopting AI technology [49]. At the moment, the BERT-based models provide a more reliable, trustworthy, resource-effective solution than other GPT-based models to leverage large amounts of useful information buried in colloquial texts such as patient narratives.

Third, these design guidelines may potentially be adapted to other domains. While our study focuses on symptom recognition for healthcare-related applications involving informal texts such as patient narratives and emails, the findings may be generalizable to other application domains where there are similar distinctions between the formal and colloquial references to concepts. For example, customer complaints about software flaws or product defects may contain colloquial descriptions of issues and user experiences. Automatic recognition of the relevant entities can help correctly categorize the problems and direct them to the appropriate care teams in a timely manner, thereby improving customer relationships.

*6.4 Limitations and Future Work*

While NER in general involves multiple entity types simultaneously, we focus in this study on a single entity class, symptom, in order to limit the potential impact of the strategies on model performance and better isolate the causes. In addition, our study only examines entity recognition in the healthcare domain without testing the training strategies in other application domains. Moreover, we have examined only six BERT-based models and their performance when used for symptom recognition. It is not certain whether the findings and training strategies generalize to fine-tuning of other pretrained Transformer models.

Our future work will seek to verify if the results can be generalized to other domains where colloquial entities are useful. We would also like to explore the impact of term versus context in which it appears on learning to predict colloquial symptoms in both generic BERT-based models and domain-specific models (e.g., BioBERT).

7. **Conclusions**

Motivated by the challenges of extraction of entities from colloquial text, this research focuses on symptom recognition from patient narratives. We seek to address the key research question: *how to leverage BERT-based models to recognize symptoms effectively from non-formal, colloquial writings*. Using the design science methodology, through experimentation we have found how patterns of model performance on colloquial data are affected by different design choices, and formulated design principles based on these findings. Our research contributes to both the NER and healthcare information systems literature and practice.

# Appendix

**Table 1.** Complete performance metrics including precision (P), recall (R), and F1 scores based on exact match. Bold face marks highest value per column per test, underlined italics – second highest.

| Test Set | Training Set | BERT | | | RoBERTa | | | DeBERTa | | | DistilBERT | | | BioBERT | | | Bio-eNER | | |
|---|---|---|---|---|---|---|---|---|---|---|---|---|---|---|---|---|---|---|---|
| | | P | R | F1 | P | R | F1 | P | R | F1 | P | R | F1 | P | R | F1 | P | R | F1 |
| **CORD-Test** | CLQ | 0.54 | 0.65 | 0.59 | 0.76 | _0.79_ | 0.77 | 0.78 | 0.72 | 0.75 | 0.60 | 0.65 | 0.62 | 0.73 | **0.84** | 0.78 | 0.69 | 0.78 | 0.74 |
| | CLQ-N | 0.63 | 0.67 | 0.65 | _0.80_ | _0.79_ | _0.79_ | 0.80 | **0.77** | **0.79** | 0.63 | 0.70 | 0.67 | 0.74 | **0.84** | _0.79_ | 0.75 | 0.79 | 0.77 |
| | MIX-D | 0.74 | 0.69 | 0.72 | **0.81** | 0.75 | 0.78 | **0.83** | 0.72 | 0.77 | **0.79** | 0.73 | 0.76 | 0.78 | 0.77 | 0.78 | 0.78 | 0.74 | 0.76 |
| | MIX | _0.75_ | **0.75** | _0.75_ | 0.79 | _0.79_ | _0.79_ | 0.80 | _0.76_ | _0.78_ | 0.73 | 0.76 | 0.75 | _0.82_ | _0.8_ | **0.81** | 0.78 | **0.82** | **0.80** |
| | MIX-N | 0.73 | _0.74_ | 0.74 | **0.81** | _0.79_ | 0.80 | 0.80 | **0.77** | **0.79** | 0.74 | 0.76 | 0.75 | **0.83** | _0.8_ | **0.81** | 0.81 | 0.79 | **0.80** |
| | CORD-D | _0.75_ | 0.72 | 0.74 | **0.81** | _0.79_ | 0.80 | _0.82_ | 0.74 | _0.78_ | 0.78 | 0.76 | **0.77** | 0.81 | 0.76 | 0.78 | 0.80 | 0.77 | 0.79 |
| | CORD | **0.78** | _0.74_ | **0.76** | _0.80_ | **0.80** | **0.80** | 0.81 | **0.77** | **0.79** | 0.77 | **0.77** | **0.77** | 0.8 | 0.78 | _0.79_ | 0.81 | 0.79 | **0.80** |
| **CLQ-Test** | CLQ | 0.48 | 0.66 | 0.56 | **0.78** | **0.87** | **0.82** | 0.77 | **0.87** | **0.81** | 0.58 | **0.70** | 0.63 | 0.68 | _0.81_ | 0.74 | 0.69 | 0.76 | 0.72 |
| | CLQ-N | 0.5 | 0.6 | 0.55 | _0.75_ | 0.76 | _0.75_ | **0.77** | 0.77 | _0.77_ | 0.57 | 0.63 | 0.6 | 0.75 | 0.8 | **0.78** | _0.74_ | 0.73 | 0.73 |
| | MIX-D | 0.59 | _0.69_ | _0.64_ | 0.73 | 0.78 | _0.75_ | _0.74_ | 0.80 | _0.77_ | 0.63 | **0.70** | _0.66_ | 0.73 | **0.84** | 0.78 | 0.68 | 0.77 | 0.72 |
| | MIX | 0.58 | **0.7** | 0.63 | 0.71 | _0.80_ | _0.75_ | 0.71 | _0.81_ | 0.76 | 0.61 | **0.70** | 0.65 | _0.76_ | 0.77 | _0.76_ | 0.7 | **0.78** | _0.74_ |
| | MIX-N | _0.6_ | 0.63 | 0.61 | 0.74 | 0.76 | _0.75_ | 0.73 | 0.76 | 0.75 | **0.69** | _0.69_ | **0.69** | 0.75 | 0.71 | 0.73 | **0.78** | 0.74 | **0.76** |
| | CORD-D | 0.49 | **0.7** | 0.58 | 0.66 | _0.80_ | 0.72 | 0.67 | 0.77 | 0.71 | 0.46 | 0.67 | 0.55 | 0.65 | 0.78 | 0.71 | 0.52 | _0.77_ | 0.62 |
| | CORD | **0.68** | 0.64 | **0.66** | 0.70 | 0.73 | 0.71 | 0.72 | 0.72 | 0.72 | _0.66_ | 0.62 | 0.64 | **0.78** | 0.7 | 0.74 | 0.72 | 0.73 | 0.72 |
| **Med Help** | CLQ | 0.45 | 0.62 | 0.52 | 0.63 | **0.75** | 0.69 | _0.67_ | **0.75** | **0.71** | 0.46 | 0.61 | 0.52 | 0.54 | **0.71** | _0.62_ | 0.53 | _0.67_ | 0.59 |
| | CLQ-N | 0.49 | 0.58 | 0.53 | _0.65_ | 0.70 | 0.68 | **0.68** | 0.71 | 0.70 | 0.50 | 0.60 | 0.55 | 0.59 | 0.68 | **0.63** | _0.61_ | _0.67_ | **0.64** |
| | MIX-D | _0.53_ | **0.65** | **0.58** | 0.64 | 0.70 | 0.67 | _0.67_ | 0.69 | 0.68 | _0.55_ | 0.64 | **0.59** | 0.56 | _0.69_ | _0.62_ | 0.57 | _0.67_ | 0.61 |
| | MIX | 0.5 | _0.63_ | 0.56 | 0.63 | _0.72_ | 0.67 | 0.66 | _0.72_ | 0.69 | 0.52 | **0.65** | 0.58 | 0.59 | 0.67 | **0.63** | 0.54 | **0.69** | 0.60 |
| | MIX-N | 0.5 | 0.6 | 0.55 | **0.66** | 0.71 | 0.68 | _0.67_ | 0.70 | 0.68 | **0.56** | 0.61 | **0.59** | _0.61_ | 0.63 | _0.62_ | 0.63 | 0.65 | **0.64** |
| | CORD-D | 0.36 | 0.62 | 0.46 | 0.52 | 0.69 | 0.59 | 0.52 | 0.68 | 0.59 | 0.33 | 0.64 | 0.44 | 0.43 | 0.65 | 0.52 | 0.33 | _0.67_ | 0.45 |
| | CORD | **0.57** | 0.58 | _0.57_ | 0.61 | 0.68 | 0.64 | 0.62 | 0.65 | 0.64 | **0.56** | 0.59 | 0.57 | _0.6_ | 0.63 | _0.62_ | _0.61_ | 0.65 | _0.63_ |
| **iCliniq** | CLQ | 0.44 | 0.61 | 0.51 | _0.61_ | **0.77** | 0.68 | **0.64** | **0.76** | **0.70** | 0.47 | 0.63 | 0.54 | 0.56 | **0.73** | _0.64_ | 0.52 | 0.69 | 0.59 |
| | CLQ-N | 0.47 | 0.58 | 0.52 | 0.59 | 0.71 | 0.65 | _0.61_ | 0.71 | 0.66 | 0.50 | 0.63 | 0.56 | 0.58 | _0.7_ | _0.64_ | 0.58 | 0.68 | 0.62 |
| | MIX-D | _0.55_ | **0.66** | 0.6 | **0.63** | 0.71 | _0.67_ | **0.64** | 0.72 | 0.68 | _0.54_ | **0.67** | **0.60** | 0.6 | _0.7_ | **0.65** | _0.59_ | **0.73** | **0.65** |
| | MIX | 0.52 | 0.63 | 0.57 | 0.59 | _0.73_ | 0.65 | _0.61_ | 0.71 | 0.66 | 0.51 | **0.67** | 0.58 | _0.61_ | 0.67 | _0.64_ | 0.54 | 0.70 | 0.61 |
| | MIX-N | 0.5 | 0.61 | 0.55 | 0.60 | 0.70 | 0.65 | _0.61_ | 0.69 | 0.65 | **0.55** | _0.64_ | _0.59_ | **0.62** | 0.64 | 0.63 | **0.61** | 0.66 | _0.64_ |
| | CORD-D | 0.46 | _0.65_ | 0.54 | 0.55 | 0.71 | 0.62 | 0.57 | 0.71 | 0.63 | 0.44 | **0.67** | 0.54 | 0.55 | 0.66 | 0.6 | 0.47 | _0.71_ | 0.57 |
| | CORD | **0.58** | 0.6 | _0.59_ | 0.57 | 0.67 | 0.62 | 0.56 | 0.66 | 0.61 | _0.54_ | 0.62 | 0.58 | 0.6 | 0.63 | 0.62 | **0.61** | 0.65 | 0.63 |